\pdfoutput=1
\documentclass[11pt,a4paper]{article}

\usepackage[utf8]{inputenc}
\usepackage[T1]{fontenc}
\usepackage[margin=1in]{geometry} 
\usepackage{amsmath,amsfonts,amssymb}
\usepackage{graphicx}
\usepackage{makecell}
\usepackage{authblk} 
\usepackage{hyperref}
\usepackage{float}

\title{\textbf{Unsupervised Metal Artifact Reduction in Dental CBCT using Fine-tuned Cycle-Consistent Adversarial Networks}}

\author[1]{G.L.T. Chamika}
\author[1]{S.N.A. Dhanapala}
\author[1]{P.H.S.V. Nimalaweera}
\author[1,*]{Maheshi B. Dissanayake}
\author[2]{Ruwan D. Jayasinghe}

\affil[1]{Department of Electrical and Electronic Engineering, University of Peradeniya, Peradeniya, Sri Lanka}
\affil[2]{Department of Oral Medicine and Periodontology, Faculty of Dental Sciences, University of Peradeniya, Peradeniya, Sri Lanka}
\affil[*]{Corresponding author email: maheshid@eng.pdn.ac.lk}

\date{\today}

\begin{document}

\maketitle


\begin{abstract}
Metal artifacts generated by dental implants significantly degrade cone-beam computed tomography (CBCT) volumes, obscuring critical anatomical structures and compromising diagnostic precision. To address this, an unsupervised deep learning framework has been proposed for Metal Artifact Reduction (MAR) utilizing a Cycle-Consistent Adversarial Network (CycleGAN) optimized for high-fidelity restoration. Unlike supervised methods that rely on unattainable voxel-aligned paired datasets, the proposed approach leverages an unpaired dataset of approximately 4,000 images, curated from the public ToothFairy dataset. The architecture integrates U-Net-based generators and PatchGAN discriminators, specifically tuned to mitigate generative hallucinations and preserve morphological integrity. Quantitative benchmarking on a held-out test set demonstrates a 34.6\% improvement in the Blind/Referenceless Image Spatial Quality Evaluator (BRISQUE) score, a substantial reduction in Fréchet Inception Distance (FID) from 207.03 to 157.04, and a superior Structural Similarity Index Measure (SSIM) of 0.9105. 
The framework achieves real-time efficiency with a 3.03 ms inference time per slice, effectively suppressing artifacts while preserving anatomical detail. Expert validation confirms high fidelity; however, to ensure reliability in extreme cases, the architecture is recommended as a clinical decision-support tool under human-in-the-loop oversight. By enhancing diagnostic clarity via a scalable software pipeline, this study provides a robust solution for high-fidelity dental implant imaging.
\end{abstract}

\noindent \textbf{Keywords:} Cone-Beam Computed Tomography; metal artifact reduction; CycleGAN; unsupervised learning; dental imaging; deep learning; image-to-image translation 

\section{Introduction}

Cone-beam computed tomography (CBCT) has become a cornerstone of the digital transformation in dental and maxillofacial medicine, providing high-resolution three-dimensional data essential for modern computer-aided diagnostics \cite{ref1,ref2}. In the era of Digital Health, CBCT data is the primary input for precision workflows in orthodontics, endodontics, and implant planning, where the accuracy of the digital twin---the virtual representation of the patient's anatomy---is critical for successful treatment outcomes \cite{ref3,ref5,ref6}. However, the reliability of this digital diagnostic pipeline is frequently compromised by metallic objects, such as dental implants and restorations. These objects introduce severe streak artifacts that degrade image data integrity, leading to potential errors in automated planning software and digital surgical guides. 
These Metallic objects create artifacts in CBCT images as a result of beam hardening, photon starvation, and scattered radiation during data acquisition \cite{ref3,ref4,ref14,ref15}. These artifacts manifest as streaking, shading, and intensity distortions that obscure anatomical details and reduce contrast resolution. In dental imaging, where metallic objects are frequently located near clinically important regions, metal artifacts can mask pathological findings or mimic disease, thereby reducing diagnostic accuracy and clinical confidence \cite{ref5,ref6,ref18}.

As of literature, a wide range of metal artifact reduction (MAR) techniques has been proposed to address this. Early approaches primarily relied on image post-processing and classical reconstruction-based strategies \cite{ref1,ref2,ref7}. Projection-domain and iterative reconstruction methods demonstrated improved artifact suppression but were sensitive to metal segmentation accuracy and often required high computational resources, limiting their routine clinical applicability \cite{ref7,ref14,ref15,ref16}.
More recently, deep learning-based MAR approaches have gained increasing attention due to their ability to model complex artifact patterns directly from data.\cite{ref28,ref29,ref30,ref31} Supervised convolutional neural networks, including U-Net-based architectures, have shown promising performance in CT and CBCT artifact reduction \cite{ref10,ref12,ref19}. Supervised methods require paired training sets of artifact-affected and artifact-free images from the same patient. Such datasets are rarely attainable in clinical practice due to stringent ethical constraints concerning dual radiation exposure and the technical difficulty of ensuring perfect anatomical registration between successive acquisitions \cite{ref10,ref12}.
Unsupervised learning frameworks provide a practical alternative by eliminating the need for paired training data. Generative adversarial networks (GANs), particularly CycleGAN, enable unpaired image-to-image translation through cycle-consistency constraints and have been successfully applied to metal artifact reduction tasks \cite{ref11,ref20,ref21}. Despite their effectiveness, GAN-based MAR methods may introduce structural hallucinations or distort fine anatomical details if not carefully regularized \cite{ref10,ref21}.

\subsection{Research Gap and Motivation}

Despite significant advancements in metal artifact reduction (MAR) for dental CBCT imaging, several critical limitations persist in existing methodologies. These limitations highlight the need for more robust, clinically viable solutions.

The key research gaps identified in the literature are as follows:

\begin{enumerate}
    \item \textbf{Dependence on Paired Training Data:}
    Supervised deep learning approaches require voxel-aligned artifact-free and artifact-affected image pairs. However, such datasets are practically unattainable in clinical settings due to ethical constraints related to repeated radiation exposure and challenges in achieving perfect anatomical registration.

    \item \textbf{Sensitivity of Classical MAR Methods:}
    Traditional projection-domain and reconstruction-based techniques rely heavily on accurate metal segmentation and simplified physical models. These approaches often fail under severe artifact conditions and may introduce secondary distortions or excessive smoothing of anatomical structures.

    \item \textbf{Risk of Structural Hallucinations in GAN-based Methods:}  
    While unsupervised GAN frameworks eliminate the need for paired data, they may generate unrealistic textures or distort fine anatomical details if not properly constrained, limiting their reliability for clinical use.

    \item \textbf{Limited Practical Integration into Clinical Workflows:}  
    Many existing MAR solutions require hardware modifications, computationally intensive reconstruction, or manual preprocessing steps, reducing their feasibility for routine deployment in digital dentistry environments.
\end{enumerate}

\textbf{{Motivation:} } 
These gaps collectively indicate the need for a data-efficient, unsupervised MAR framework that can operate without paired datasets, while preserving anatomical fidelity, minimizing artificial distortions, and integrating seamlessly into real-world clinical workflows. This study is motivated by the objective of addressing these limitations through a carefully regularized CycleGAN-based approach.

\subsection{Study Overview and Contributions}
In this study, we investigate a fine-tuned unsupervised CycleGAN-based metal artifact reduction framework for dental CBCT imaging. The proposed method leverages unpaired artifact-affected and artifact-free datasets and employs U-Net-based generators with PatchGAN discriminators and carefully optimized loss functions to suppress metal artifacts while preserving anatomically relevant structures. By avoiding the need for paired data, explicit metal segmentation, or hardware modifications, the proposed approach aligns with real-world clinical constraints and enhances diagnostic clarity in dental CBCT imaging \cite{ref10,ref11,ref13,ref20}.

The primary contributions of this research, aligned with the identified research gaps, are as follows:

\begin{enumerate}
 
   \item\textbf{Addressing the Lack of Paired Data:}  
  We implement an unsupervised CycleGAN-based MAR pipeline that eliminates the need for voxel-perfect paired datasets, directly addressing the primary limitation of supervised approaches.

  \item \textbf{Improved Anatomical Preservation in GAN Frameworks:}  
  By integrating U-Net-based generators with PatchGAN discriminators and carefully tuned loss functions, the proposed model mitigates structural hallucinations while preserving fine anatomical details.

  \item \textbf{Robustness Beyond Classical MAR Limitations:}  
  Unlike traditional methods that depend on metal segmentation or physical modeling, the proposed approach learns artifact distributions directly from data, improving performance under complex artifact conditions.

  \item \textbf{Practical Workflow Integration:}  
  The proposed solution does not require hardware modifications or manual preprocessing, making it suitable for scalable deployment in existing digital dentistry pipelines.

\end{enumerate}

The remainder of this paper is structured as follows. Section 2 presents the background and reviews existing metal artifact reduction techniques, including classical and deep learning-based approaches, along with identified challenges. Section 3 describes the proposed methodology, including dataset preparation, CycleGAN architecture, and training strategy. Section 4 presents the experimental results, including quantitative and qualitative evaluations. Section 5 discusses the findings, limitations, and clinical implications of the proposed approach. Finally, Section 6 concludes the paper and outlines potential directions for future research.


\section{Background}

\subsection{Research Problem}
Cone-beam computed tomography (CBCT) has been widely adopted in maxillofacial diagnostics because low-dose three-dimensional visualization of hard tissues can be achieved \cite{ref1,ref2}. Nevertheless, its clinical utility is frequently compromised by the presence of high-density metallic restorations and implants. Severe artifacts, commonly manifested as streaking and shading, are introduced by physical effects such as beam hardening, photon starvation, and scatter radiation \cite{ref3,ref4}.

In dental CBCT imaging, metallic objects are often located in close proximity to anatomically and diagnostically critical regions, including the alveolar bone, root canals, and periodontal structures. As a result, fine anatomical details may be obscured, contrast resolution may be reduced, and diagnostic accuracy in orthodontics, endodontics, and implant planning may be adversely affected \cite{ref5,ref6}. In certain cases, artifact patterns may resemble pathological findings or may conceal clinically relevant abnormalities, thereby increasing the risk of misdiagnosis and suboptimal treatment planning.

Initial attempts at metal artifact reduction were largely based on image-domain post-processing techniques. A multi-stage pipeline incorporating Gaussian Mixture Model (GMM) based tooth extraction and morphological filtering was proposed by Johari et al. \cite{ref2}, through which partial artifact suppression was achieved. However, sensitivity to threshold selection and anatomical variability was also observed. Similarly, improvements in three-dimensional visualization were reported by Ibraheem \cite{ref1} using image processing techniques, but limited robustness under complex or severe artifact conditions was noted. These early limitations indicated that more advanced MAR strategies were required for diverse clinical dental CBCT scenarios.

\subsection{Traditional and Classical MAR Methods}
Traditional metal artifact reduction (MAR) techniques have generally been categorized into projection-domain, reconstruction-domain, and image-domain approaches. In projection-domain methods, corrupted sinogram regions are corrected after metal traces have been identified, and affected projections are replaced through interpolation or inpainting strategies. Among these methods, normalized metal artifact reduction (NMAR) has been widely investigated, where projection data are normalized prior to reconstruction so that beam hardening effects can be compensated and streak artifacts can be reduced \cite{ref14}. However, strong dependence on accurate metal segmentation has been reported, and segmentation errors may lead to secondary artifacts or excessive smoothing of anatomical structures \cite{ref15}.

In reconstruction-domain methods, physical models and prior information are incorporated directly into the reconstruction process. Iterative reconstruction frameworks, including statistical and maximum-a-posteriori formulations, have been shown to provide better artifact suppression than conventional analytical reconstruction methods \cite{ref7}. A projection completion strategy combined with iterative reconstruction was presented by Lemmens et al. \cite{ref7}, and notable streak reduction was achieved. However, these methods are generally associated with high computational cost, sensitivity to parameter selection, and limited practicality in routine dental CBCT workflows. Weighted filtered back-projection and hybrid reconstruction schemes have also been explored, but their performance has been shown to deteriorate under severe attenuation conditions that are frequently encountered in dental imaging \cite{ref16}.

Additional challenges are introduced in dental CBCT systems by variations in scanner geometry, limited projection angles, and low-dose acquisition protocols. As a result, the robustness and generalizability of classical MAR methods across different clinical environments are often reduced. Moreover, handcrafted assumptions and simplified physical models are typically relied upon, and these are often insufficient to represent the nonlinear and spatially varying characteristics of metal-induced artifacts in real dental CBCT images \cite{ref6}.

Overall, meaningful artifact suppression has been achieved by traditional and classical MAR approaches, particularly under controlled conditions and moderate artifact severity. However, limited robustness, sensitivity to segmentation and acquisition conditions, and high computational burden have remained major constraints on their routine clinical applicability.

\subsection{Deep Learning Approaches for MAR}
With the rapid advancement of deep learning, data-driven alternatives to conventional MAR methods have been introduced, allowing complex artifact patterns to be learned directly from imaging data. Supervised convolutional neural networks (CNNs), particularly U-Net-based architectures, have been widely applied to CT and CBCT artifact reduction by learning mappings between artifact-corrupted images and corresponding artifact-free references \cite{ref10,ref12,ref19}. Superior performance relative to many conventional methods has been reported. However, clinical applicability in dental CBCT has been limited by the requirement for paired training data, since metal-free ground-truth images from the same patient and acquisition setting are rarely obtainable in practice and may be ethically difficult to acquire.

To overcome this limitation, unsupervised learning frameworks have been explored. In particular, generative adversarial networks (GANs), especially CycleGAN, have enabled unpaired image-to-image translation through the enforcement of cycle consistency between source and target domains \cite{ref11}. Effective reduction of metal artifacts in CT images using unpaired data was demonstrated by Du et al. \cite{ref11}, indicating that CycleGAN-based methods are suitable for MAR under realistic clinical constraints. Nevertheless, early GAN-based MAR methods were also found to be vulnerable to unrealistic texture generation and structural hallucinations.

Subsequent studies have therefore focused on improving structural fidelity and training stability through architectural refinement and loss-function design. An attention-guided $\beta$-CycleGAN framework was introduced by Lee et al. \cite{ref10}, through which anatomical preservation during artifact reduction was improved. More recent GAN-based and hybrid learning strategies have been extended to dental CBCT imaging, and improved artifact suppression with preservation of clinically relevant anatomical structures has been reported \cite{ref20,ref21,ref22,ref23,ref27}. In parallel, emerging generative paradigms, including diffusion-based models and transformer-based architectures, have shown promise for modeling complex artifact distributions and improving restoration quality in volumetric CBCT imaging \cite{ref24,ref25,ref26,ref28,ref31}.


Recent advancements in deep learning-based restoration have demonstrated significant improvements in artifact suppression by leveraging sophisticated priors and multi-modal fusion. For instance, recent paradigms such as WaterCycleDiffusion utilize visual-textual fusion and stable diffusion priors to enhance underwater imagery \cite{ref33}, while mutual guidance fusion networks have been successfully applied to remote sensing to improve visual reasoning \cite{ref32}. These cross-disciplinary successes underscore the efficacy of using generative priors to recover information from highly corrupted visual data. However, while these advanced architectures often rely on auxiliary textual prompts or paired reference data—inputs that are fundamentally unavailable in clinical dental CBCT—our proposed framework achieves comparable structural recovery through an unsupervised, self-consistent learning strategy.

In summary, the primary limitations in this domain remain the heavy dependence on voxel-aligned paired data for supervised training and the inherent risk of structural hallucinations in unsupervised models. Furthermore, existing frameworks often fail to generalize across diverse metallic prosthetics or different CBCT hardware vendors. There is a critical need for robust, unsupervised strategies that maintain high anatomical fidelity—verified by clinical expertise—while remaining adaptable to the varied artifact signatures found in real-world institutional data.

\subsection{Challenges and Research Gaps}
Despite substantial progress, several challenges remain unresolved in MAR for dental CBCT imaging. Supervised deep learning methods continue to be constrained by the scarcity of paired datasets, whereas traditional MAR techniques remain sensitive to segmentation errors and frequently show degraded performance under severe artifact conditions \cite{ref10,ref14}. Although unsupervised GAN-based models alleviate data pairing requirements, insufficient regularization may lead to hallucinated structures or subtle distortion of anatomical features, which is unacceptable in clinical contexts \cite{ref21,ref23,ref29,ref30}.

Computational efficiency is another critical concern, particularly for iterative reconstruction and high-capacity deep learning models, which may hinder real-time clinical deployment \cite{ref29,ref30}. In addition, performance evaluation in dental CBCT MAR remains challenging because standardized benchmarks and reliable artifact-free ground truth are generally unavailable. Although quantitative measures such as PSNR and SSIM have been widely used, diagnostic relevance and clinical interpretability are not always adequately reflected by these metrics \cite{ref6,ref22}.

These unresolved issues indicate that MAR solutions that can operate effectively on unpaired data, preserve anatomical fidelity, minimize unrealistic corrections, and remain computationally practical for routine clinical use, are in need  \cite{ref10,ref14,ref21,ref23}. In this context, CycleGAN-based frameworks have been regarded as particularly promising because unpaired image translation can be achieved through adversarial and cycle-consistent learning \cite{ref10,ref11,ref20}. Through the cycle-consistency constraint, structural preservation can be encouraged by ensuring that translated images can be mapped back to their original domain, thereby reducing the likelihood of hallucinated or distorted anatomical features \cite{ref10,ref11,ref21}. Furthermore, by incorporating an identity loss, unnecessary modifications to images that already resemble the target domain can be discouraged, thereby preserving clinically relevant intensity patterns \cite{ref10,ref11,ref21}.

From a computational standpoint, the direct image-to-image mapping of CycleGAN architectures eliminates the overhead of iterative optimization and complex physical modeling, providing a more efficient alternative to conventional metal artifact reduction techniques \cite{ref7,ref14,ref16}. When lightweight U-Net generators and patch-based discriminators are employed, a favorable balance between representational capacity and computational cost can be achieved \cite{ref8,ref10,ref11,ref12}. These properties provided the motivation for the adoption of a CycleGAN-based strategy in this study as a practical means of addressing the major limitations of current dental CBCT metal artifact reduction approaches \cite{ref10,ref20,ref21,ref23}.

\subsection{Contribution}
Motivated by the identified limitations, a fine-tuned unsupervised CycleGAN-based framework for metal artifact reduction in dental CBCT imaging is proposed in this study. The adverse effects of metal artifacts are learned to be reduced through the use of unpaired artifact-affected and artifact-free images during training, thereby maintaining alignment with realistic clinical constraints where paired data are not available \cite{ref10,ref11}. Within the proposed CycleGAN, U-Net-based generators are employed for the capture of multi-scale anatomical features, whereas PatchGAN discriminators are utilized to enforce local texture realism.

During training, cycle-consistency and identity loss terms are carefully tuned so that metal-induced artifacts can be suppressed while structural hallucinations are minimized and anatomical features such as teeth and alveolar bone are preserved. In contrast to classical MAR methods, explicit metal segmentation and hardware modification are not required by the proposed framework, thereby improving adaptability across different CBCT systems. In this way, a balanced trade-off between effective artifact reduction and diagnostic integrity is sought, and the key research gaps in current dental CBCT MAR methodologies are addressed \cite{ref13,ref20,ref21,ref22,ref23,ref24,ref25,ref26}.

\section{Materials and Methods}

This study proposes an unsupervised metal artifact reduction (MAR) approach using a fine-tuned CycleGAN to mitigate streaks and glare in dental cone-beam computed tomography (CBCT) images while preserving anatomical structures, such as teeth and alveolar bone. The CycleGAN leverages unpaired image-to-image translation to map artifact-affected images to the artifact-free domain, addressing the clinical challenge of paired data scarcity \cite{ref10}. The architecture is optimized to enhance diagnostic clarity without hardware modifications.

\subsection{Dataset Creation}
The images were derived from volumetric CBCT scans in Digital Imaging and Communications in Medicine (DICOM) format and converted to PNG to facilitate efficient training. Data were curated from the publicly available ToothFairy dataset \cite{ref13}, comprising approximately 4,000 two-dimensional slices. These were partitioned into artifact-affected ($n \approx 2000$) and artifact-free ($n \approx 2000$) cohorts. As the dataset was inherently balanced upon collection, no additional class-balancing procedures were required.


\textbf{Preprocessing:}  
A series of preprocessing steps were applied to ensure consistency and stability during training. First, intensity normalization was performed using min--max scaling, whereby pixel intensities were linearly mapped to the range $[-1,1]$ to align with the Tanh activation function used in the generator output layer. This normalization ensured stable gradient propagation during adversarial training.

Subsequently, all images were resized to a spatial resolution of $256 \times 256$ pixels using bilinear interpolation. This resolution was selected as a trade-off between preserving anatomical detail and maintaining computational efficiency within GPU memory constraints.

To reduce variability arising from scanner-specific acquisition differences, histogram distribution alignment was implicitly enforced through normalization rather than explicit histogram matching, thereby preserving relative contrast relationships between anatomical structures.

\textbf{Data Augmentation:}  
To improve model generalization and reduce overfitting, data augmentation techniques were applied during training. These included random horizontal flipping, small-angle rotations (within $\pm 10^\circ$), and minor scaling transformations. Augmentation was applied probabilistically at runtime, ensuring that the model was exposed to diverse orientational variations while preserving clinical realism.

No artificial artifact simulation was introduced, as the objective was to learn artifact distributions directly from real clinical data. Overall, the preprocessing and augmentation pipeline ensured that the dataset remained balanced, standardized, and representative of real-world CBCT imaging variability.

\subsection{Proposed Algorithm}

\begin{figure}[H]
\centering
\includegraphics[width=10cm]{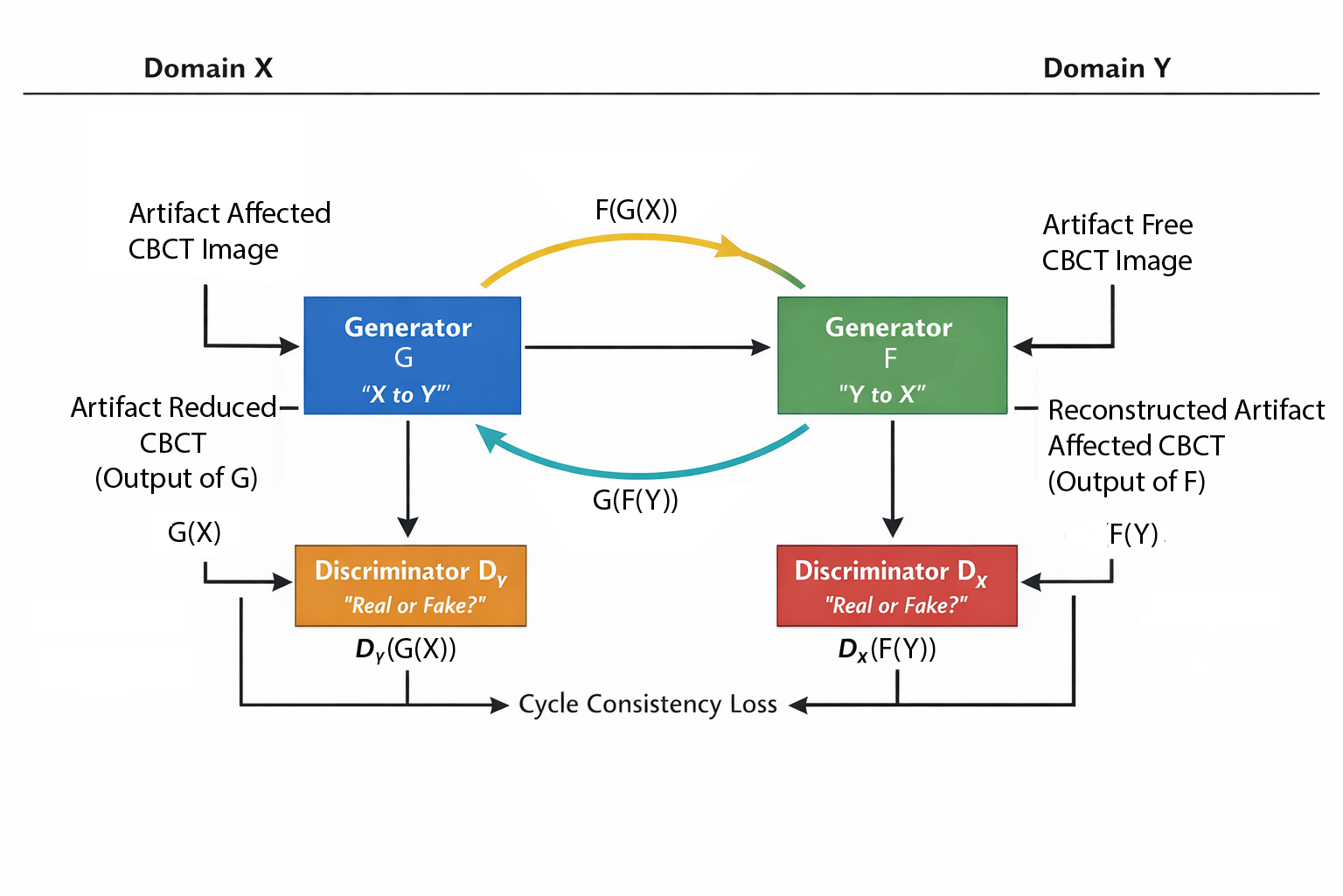}
\caption{Schematic of the proposed CycleGAN architecture for CBCT metal artifact reduction. The framework comprises two adversarial loops: (1) Forward Cycle, where Generator $G$ transforms artifact-affected images ($X$) into artifact-free reconstructions ($G(X)$), and (2) Backward Cycle, where Generator $F$ maps the clean domain ($Y$) back to the artifact-affected domain ($F(Y)$). Adversarial critics $D_Y$ and $D_X$ enforce domain realism, while the Cycle-Consistency Loss ensures anatomical fidelity by requiring that $F(G(X)) \approx X$ and $G(F(Y)) \approx Y$.}
\label{fig:discloss_cycle}
\end{figure}

The proposed CycleGAN framework, presented in Figure. \ref{fig:discloss_cycle} employs a carefully designed U-Net-inspired generator architecture comprising five encoder and five decoder layers interconnected via skip connections. This design enhances spatial feature propagation and effectively preserves fine anatomical details crucial for clinical interpretation \cite{ref8,ref10,ref11,ref12,ref19}.
In conjunction with a three-layer PatchGAN discriminator, the model learns robust bidirectional mappings between glare-affected and glare-free dental CBCT image domains by focusing on local image patches, thereby promoting realistic texture synthesis and effective artifact suppression \cite{ref10,ref11,ref12,ref20}. The network utilizes instance normalization and symmetric downsampling--upsampling stages to balance expressive feature learning while maintaining structural integrity \cite{ref10,ref12,ref19}.

\textbf{Optimization Strategy and Hyperparameter Tuning:}

The optimization strategy was designed to achieve a balance between artifact suppression and anatomical preservation. Feature extraction was implicitly performed through the hierarchical convolutional structure of the U-Net generator, enabling multi-scale representation of dental anatomical structures without the need for explicit feature selection. The final hyperparameter configuration was determined through a systematic, empirical ablation study (detailed in Section 4) to ensure optimal artifact suppression without morphological distortion, thereby justifying the optimality of the chosen loss weights beyond standard literature defaults. A coarse-to-fine tuning strategy was adopted, where initial values were selected from prior studies \cite{ref10,ref11,ref20} and subsequently refined based on validation loss convergence and visual inspection of reconstructed outputs. Particular emphasis was placed on tuning the cycle-consistency (${\lambda_\mathrm{cycle}}$) and identity (${\lambda_\mathrm{id}}$) loss weights, as these directly influence structural fidelity and the suppression of generative hallucinations. Model training was performed using the Adam optimizer with a learning rate of $0.0002$ and momentum parameters $\beta_1 = 0.5$ and $\beta_2 = 0.999$ \cite{ref9,ref10,ref11}. A summary of the hyperparameters adopted in the study is listed in the Table \ref{tab:hyperparameters}

\begin{table}[H]
\centering
\caption{Finalized hyperparameter configurations used for CycleGAN training}
\begin{tabular}{|c|c|}
\hline
\textbf{Hyperparameter} & \textbf{Value} \\
\hline
Learning rate & $2 \times 10^{-4}$ \\
\hline
Optimizer & Adam \\
\hline
$\beta_1$ & 0.5 \\
\hline
$\beta_2$ & 0.999 \\
\hline
Batch size & 1 \\
\hline
Number of epochs & 600 \\
\hline
${\lambda_\mathrm{cycle}}$ & 18 \\
\hline
${\lambda_\mathrm{identity}}$ & 15 \\
\hline
Image resolution & $256 \times 256$ \\
\hline
Normalization range & $[-1,1]$ \\
\hline
\end{tabular}
\label{tab:hyperparameters}
\end{table}


The generators are optimized using a composite loss function that integrates adversarial loss, cycle-consistency loss, and identity loss \cite{ref10,ref11,ref20,ref21}. The \textbf{cycle-consistency loss} enforces structural preservation by ensuring that an image translated to the target domain and subsequently mapped back to the original domain remains
unchanged \cite{ref10,ref11}. It is defined as in Equation~\ref{eq_cycle_loss}:

\begin{equation} \label{eq_cycle_loss}
{\mathcal{L}_\mathrm{cyc}}(G, F) =
\mathbb{E}_{x \sim X}\left[\|F(G(x)) - x\|_{{\mathrm{1}}}\right] +
\mathbb{E}_{y \sim Y}\left[\|G(F(y)) - y\|_{{\mathrm{1}}}\right]
\end{equation}

The \textbf{identity loss}, as defined in Equation~\ref{eq_identity}, constrains the generators to avoid unnecessary modifications when the input image already belongs to the target domain, thereby preserving intensity distribution and texture consistency \cite{ref10,ref11,ref21}: 

\begin{equation} \label{eq_identity}
{\mathcal{L}_\mathrm{id}}(G, F) =
\mathbb{E}_{y \sim Y}\left[\|G(y) - y\|_{{\mathrm{1}}}\right] +
\mathbb{E}_{x \sim X}\left[\|F(x) - x\|_{{\mathrm{1}}}\right]
\end{equation}

The total generator objective is defined as in Equation~\ref{eq_totalLoss}~\cite{ref10,ref11,ref20}:

\begin{equation} \label{eq_totalLoss}
{\mathcal{L}_{\mathrm{G}}^{\mathrm{total}}} =
{\mathcal{L}_{\mathrm{GAN}}}(G, {\mathbf{D}_Y}) +
{\mathcal{L}_{\mathrm{GAN}}}(F, {\mathbf{D}_X}) +
{\lambda_{\mathrm{cycle}}}\,{\mathcal{L}_{\mathrm{cyc}}} +
{\lambda_{\mathrm{identity}}}\,{\mathcal{L}_{\mathrm{id}}},
\end{equation}
where the weighting coefficients are set to ${\lambda_{\mathrm{cycle}}} = 18$ and
${\lambda_{\mathrm{identity}}} = 15$ \cite{ref10,ref11,ref20,ref21}. This integrated optimization strategy enables the U-Net generators to effectively suppress glare artifacts while minimizing distortion of clinically relevant dental anatomical structures, resulting in improved visual quality and reduced artifact-induced degradation across the processed images \cite{ref10,ref20,ref21,ref23}.

\subsubsection{Generator: U-Net} 

\begin{figure}[H]
\centering
\includegraphics[width=\linewidth]{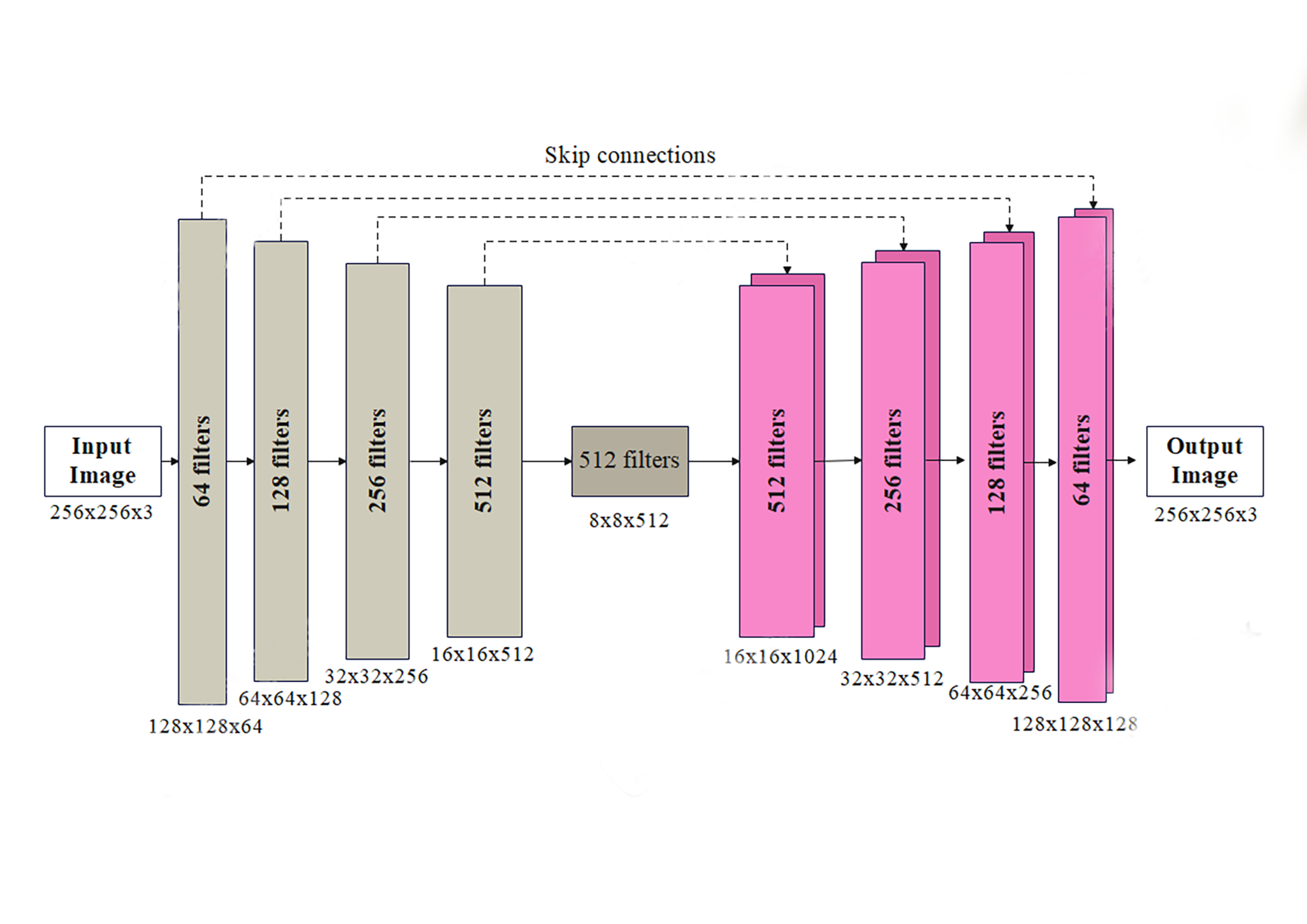}
\caption{U-Net based generator architecture.}
\label{fig:unet}
\end{figure}

The Generator follows a symmetrical U-Net architecture \cite{ref8, ref10, ref11, ref12}, comprising a 10-layer fully convolutional network structured into five downsampling (encoder) and five upsampling (decoder) stages shown in Figure~\ref{fig:unet}. This hierarchical design facilitates the extraction of multi-scale features, progressively contracting the input spatial dimensions from $256 \times 256$ to an $8 \times 8$ bottleneck. The encoder layers utilize strided convolutions to capture high-level contextual information while expanding the feature channel depth. To mitigate the loss of spatial information during the contraction phase, skip connections are implemented between corresponding encoder and decoder stages. These connections facilitate direct feature propagation, enabling the decoder to concatenate low-level structural details with high-level semantic features during transpose convolutions. This mechanism is critical for the high-fidelity recovery of fine anatomical structures, such as the periodontal ligament space and trabecular patterns, which are often obscured by metal-induced glare. Non-linearity is introduced via Rectified Linear Unit (ReLU) activations across hidden layers, while the terminal layer employs a Tanh activation to ensure output pixel values are constrained within the normalized range of $[-1, 1]$. This architecture ensures structural consistency and robust artifact suppression by effectively leveraging multi-resolution feature maps, thereby maintaining the clinical diagnostic integrity of the reconstructed dental volumes.

\subsubsection{Discriminator: PatchGAN}

\begin{figure}[H]
\centering
\includegraphics[width=12cm]{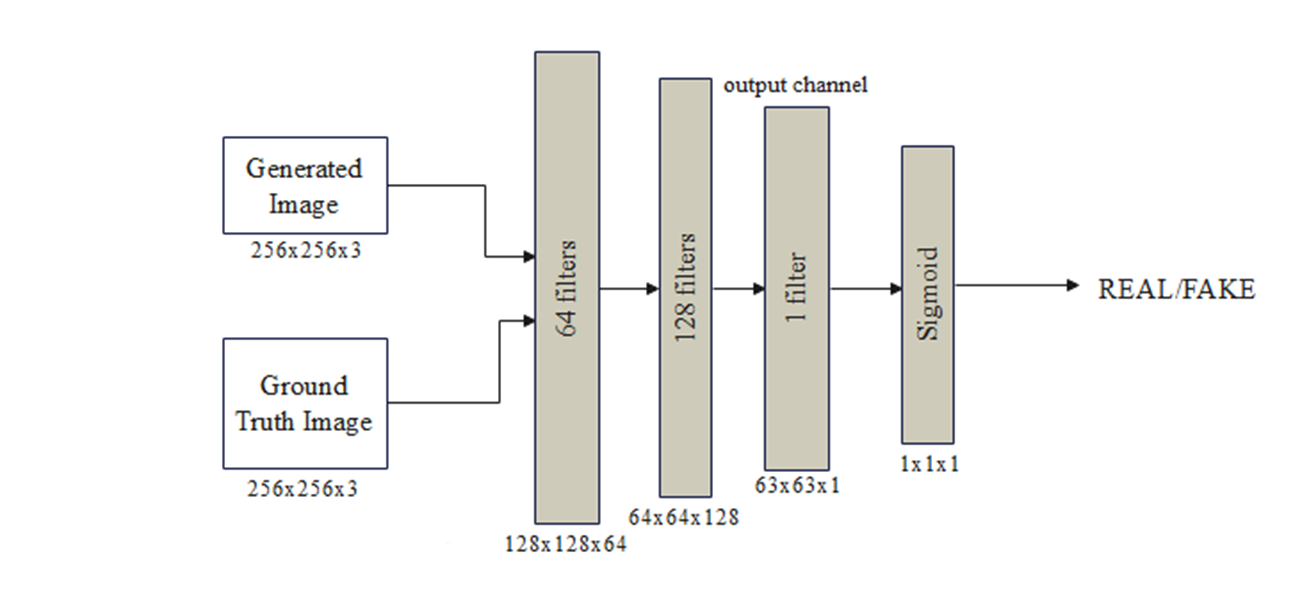}
\caption{PatchGAN discriminator architecture.}
\label{fig:patchgan}
\end{figure}

The proposed generator operates within an adversarial framework coupled with a PatchGAN discriminator \cite{ref8, ref10, ref11}, a specialized three-layer convolutional architecture ( as shown in Figure~\ref{fig:patchgan}) designed to evaluate local image realism. Unlike traditional discriminators that output a single scalar for the entire image, the PatchGAN architecture classifies $N \times N$ local patches as real or synthetic. This localized evaluation enforces high-frequency structural consistency, which is essential for suppressing fine-scale streak artifacts while preserving dental morphology. The discriminator processes $256 \times 256$ input volumes through a sequence of three convolutional layers with progressively increasing channel depths. The initial layer extracts features with 64 channels while downsampling the spatial resolution to $128 \times 128$; the second layer doubles the channel depth to 128 at a $64 \times 64$ resolution; and the terminal layer produces a single-channel feature map. This output grid represents the probability of realism across overlapping local regions. To facilitate stable adversarial training and avoid gradient vanishing, Leaky ReLU activations are applied following each convolution, with a final Sigmoid activation mapping the patch-wise predictions to the $[0, 1]$ interval. This hierarchical, patch-wise approach optimizes the model's focus on regional artifact reduction, yielding sharper, artifact-free reconstructions that maintain the diagnostic integrity required for maxillofacial radiology.

\subsection{Training, Validation, and Testing}

The dataset was partitioned into training, validation, and testing subsets using a randomized split with a ratio of 70:15:15. This partitioning strategy was selected to ensure sufficient data availability for model learning while maintaining statistically meaningful validation and testing subsets for performance evaluation.

The split was performed at the patient level using uniform random sampling, ensuring that samples from both artifact-affected and artifact-free domains were proportionally represented across all subsets. This approach preserved dataset balance within each partition.

The 70\% training subset was utilized for parameter optimization and model training, while the 15\% validation subset was employed for monitoring convergence behavior and preventing overfitting. The remaining 15\% test subset was strictly isolated and used only for final performance evaluation.

A hold-out validation strategy was adopted instead of cross-validation due to the high computational cost associated with adversarial training and the large dataset size. Given the dataset scale and diversity, the selected split was considered sufficient to ensure generalization without introducing excessive computational overhead.

Adversarial training was conducted using a stochastic batch size of one, a configuration commonly adopted in high-resolution image-to-image translation tasks to stabilize gradient updates \cite{ref10,ref11}. Optimization was performed using the Adam optimizer with a base learning rate of $2 \times 10^{-4}$ and momentum coefficients $\beta_1 = 0.5$ and $\beta_2 = 0.999$.

The network was trained for 600 epochs, after which convergence of adversarial and cycle-consistency losses was observed. The final model parameters were selected from the terminal epoch, as no evidence of overfitting or divergence was detected in validation loss trends.

\section{Results} 

\subsection{Training Dynamics and Model Convergence }

The training progression and convergence characteristics of the CycleGAN architecture were evaluated through an analysis of the adversarial and cycle-consistency loss curves \cite{ref9}. The Generator ($G$) loss was monitored to assess the model's proficiency in synthesizing artifact-free dental reconstructions capable of bypassing the adversarial critic. 
 
\begin{figure}[H]
\centering
\includegraphics{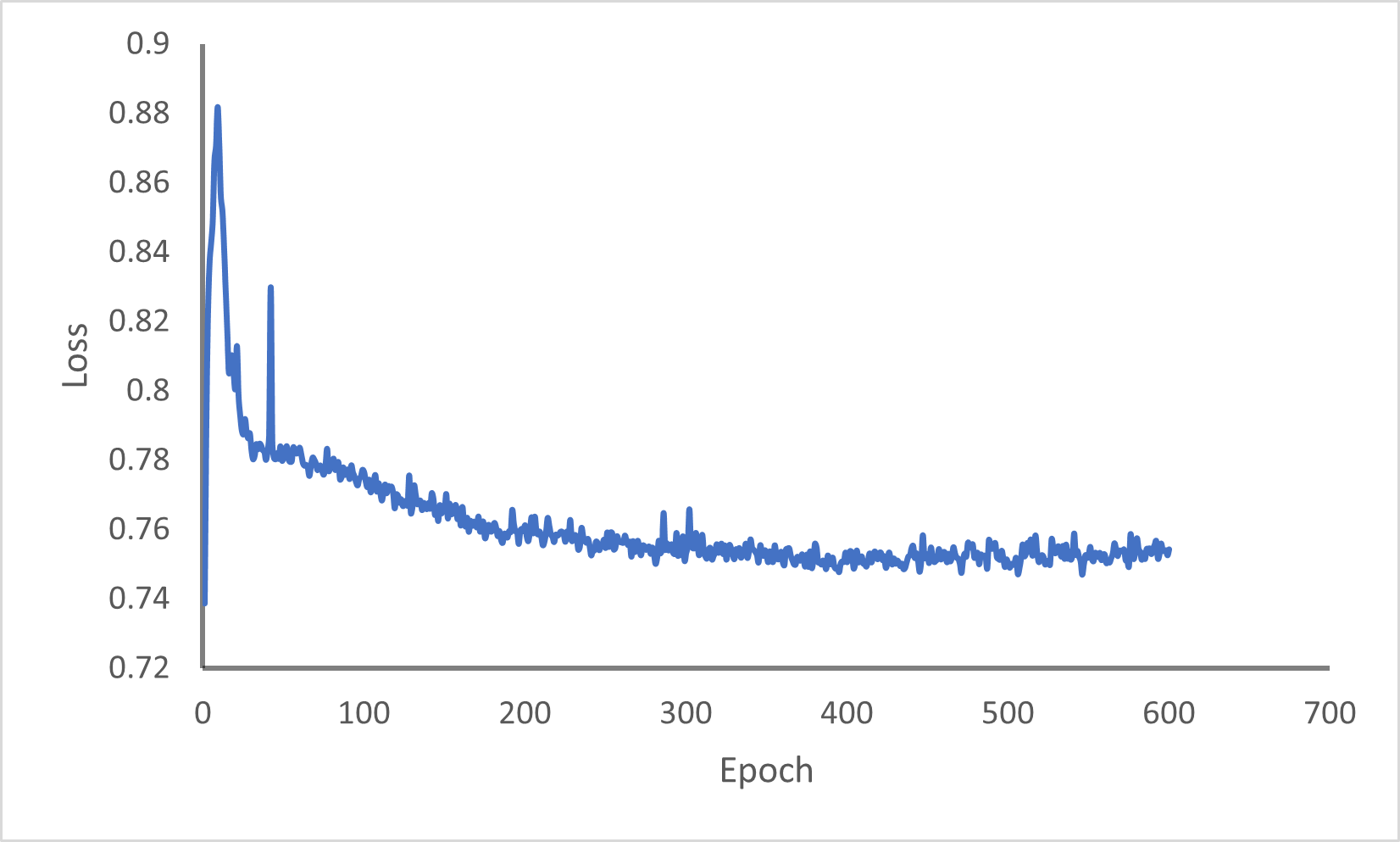}
\caption{ Generator ($G$) Loss Progression with Epoch. The trajectory illustrates the model’s proficiency in mapping artifact-affected CBCT slices to the artifact-free domain while maintaining structural consistency and adversarial realism.}
\label{fig:gen_Gloss}
\end{figure}

\begin{figure}[H]
\centering
\includegraphics{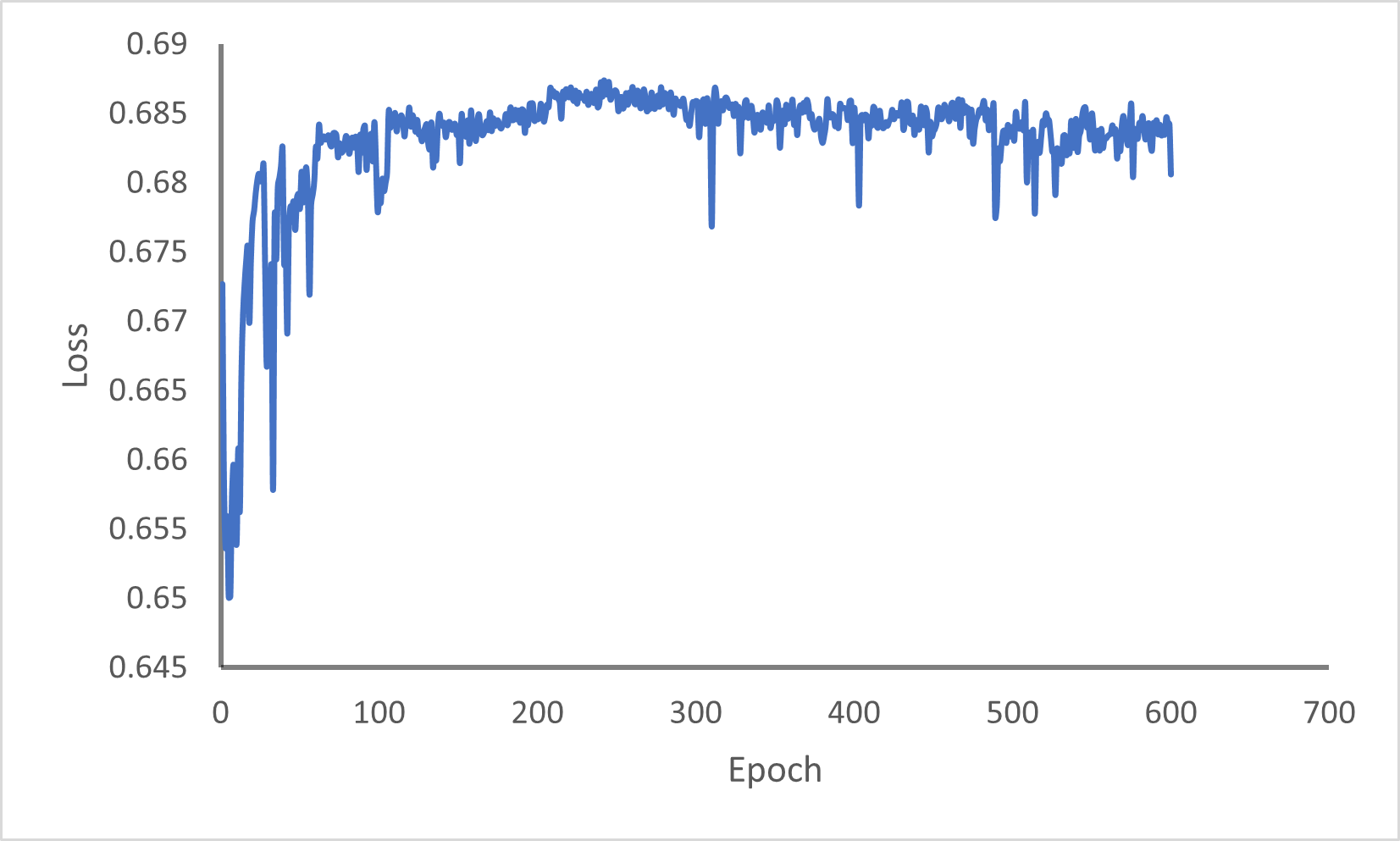}
\caption{ Discriminator ($D_Y$) Loss. The trajectory shows stable convergence within the $0.68–0.685$ range, signifying near-ideal equilibrium and the successful synthesis of artifact-free reconstructions that effectively emulate authentic clinical textures.}
\label{fig:disc_Yloss}
\end{figure}

Simultaneously, the Generator ($F$) loss ensured the fidelity of the inverse mapping: from artifact-free to artifact-affected, maintaining structural integrity through the cycle-consistency constraint. To maintain the adversarial equilibrium, the Discriminator losses (${\mathbf{D}_X}$ and ${\mathbf{D}_Y}$) were benchmarked, representing the model's ability to differentiate between authentic clinical scans and generated outputs for both artifact-affected and artifact-free domains. As illustrated in the loss trajectories in Figure \ref{fig:gen_Gloss} and \ref{fig:disc_Yloss}, the generators exhibited a progressive decline in error, while the discriminators maintained a stable, non-zero loss, indicating a well-balanced adversarial interplay. This equilibrium prevented mode collapse and ensured that the network prioritized the suppression of ``white cast'' glare without compromising the underlying diagnostic morphology. Such stability in the training phase is a critical indicator of the model's ability to generalize across varied clinical dental pathologies.

\begin{figure}[H]
\centering
\includegraphics{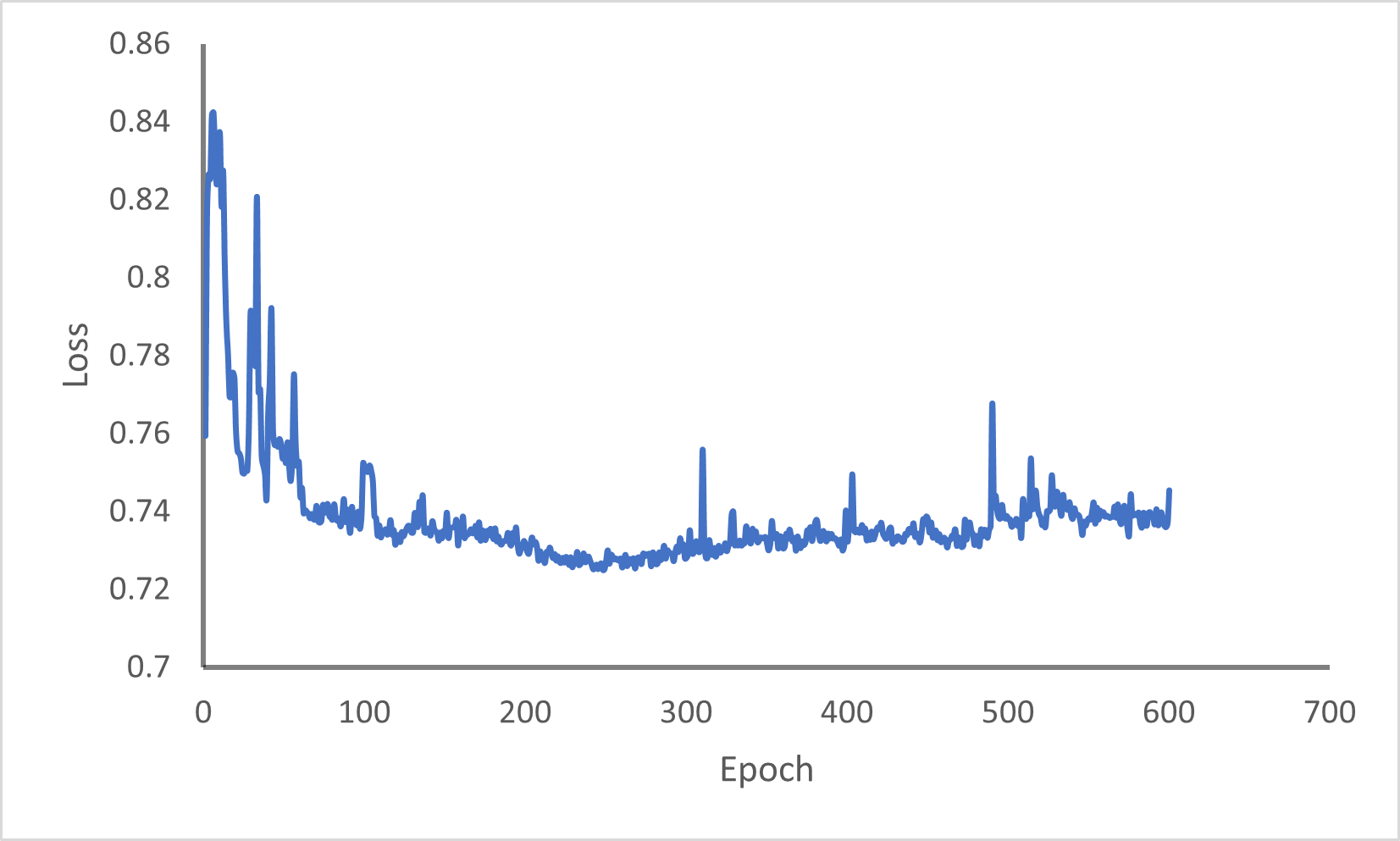}
\caption{Generator ($F$) Loss Progression with Epoch. The trajectory illustrates the inverse mapping proficiency from artifact-free to artifact-affected domains, ensuring structural integrity and cyclic stability through the cycle-consistency constraint.}
\label{fig:gen_Floss}
\end{figure}

\begin{figure}[H]
\centering
\includegraphics{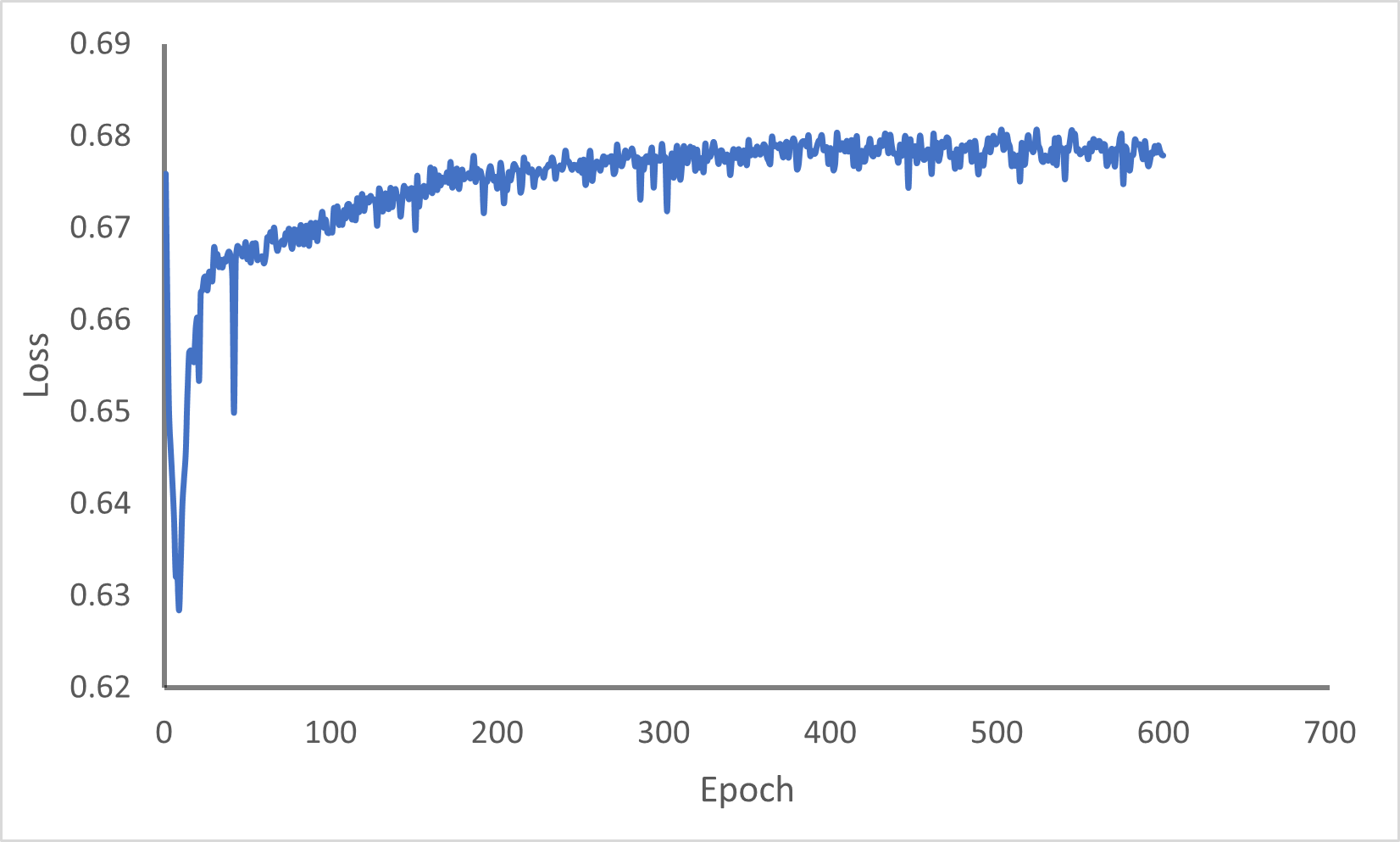}
\caption{ Discriminator ($D_X$) Loss. The trajectory demonstrates a stable convergence near the theoretical ideal ($\approx 0.69$), confirming a balanced adversarial equilibrium and the critic's proficiency in distinguishing authentic artifact-affected scans from synthesized outputs.}
\label{fig:disc_Xloss}
\end{figure}

The total objective function for Generator $G$ was formulated as a multi-component loss designed to optimize artifact suppression while safeguarding morphological integrity. This composite loss integrates the adversarial loss (${\mathcal{L}_\mathrm{adv}}$), which drives the synthesis of realistic, artifact-free reconstructions, with the cycle-consistency loss (${\mathcal{L}_\mathrm{cyc}}$) to ensure a bijective mapping between domains. The latter is particularly critical in dental imaging for maintaining the spatial constancy of fine structures like the periodontal ligament and trabecular bone. Furthermore, an identity loss (${\mathcal{L}_\mathrm{id}}$) was employed to regularize the generator, preventing unnecessary alterations to images already residing in the target domain and thereby enhancing training stability. The simultaneous minimization of these loss components allowed Generator $G$ to achieve an optimal balance between the removal of high-intensity ``white cast'' artifacts and the preservation of original diagnostic features. This multifaceted optimization strategy ensures that the generated outputs are not only visually realistic but also clinically reliable for diagnostic interpretation.

\begin{figure}[H]
\centering
\includegraphics[width=0.8\textwidth]{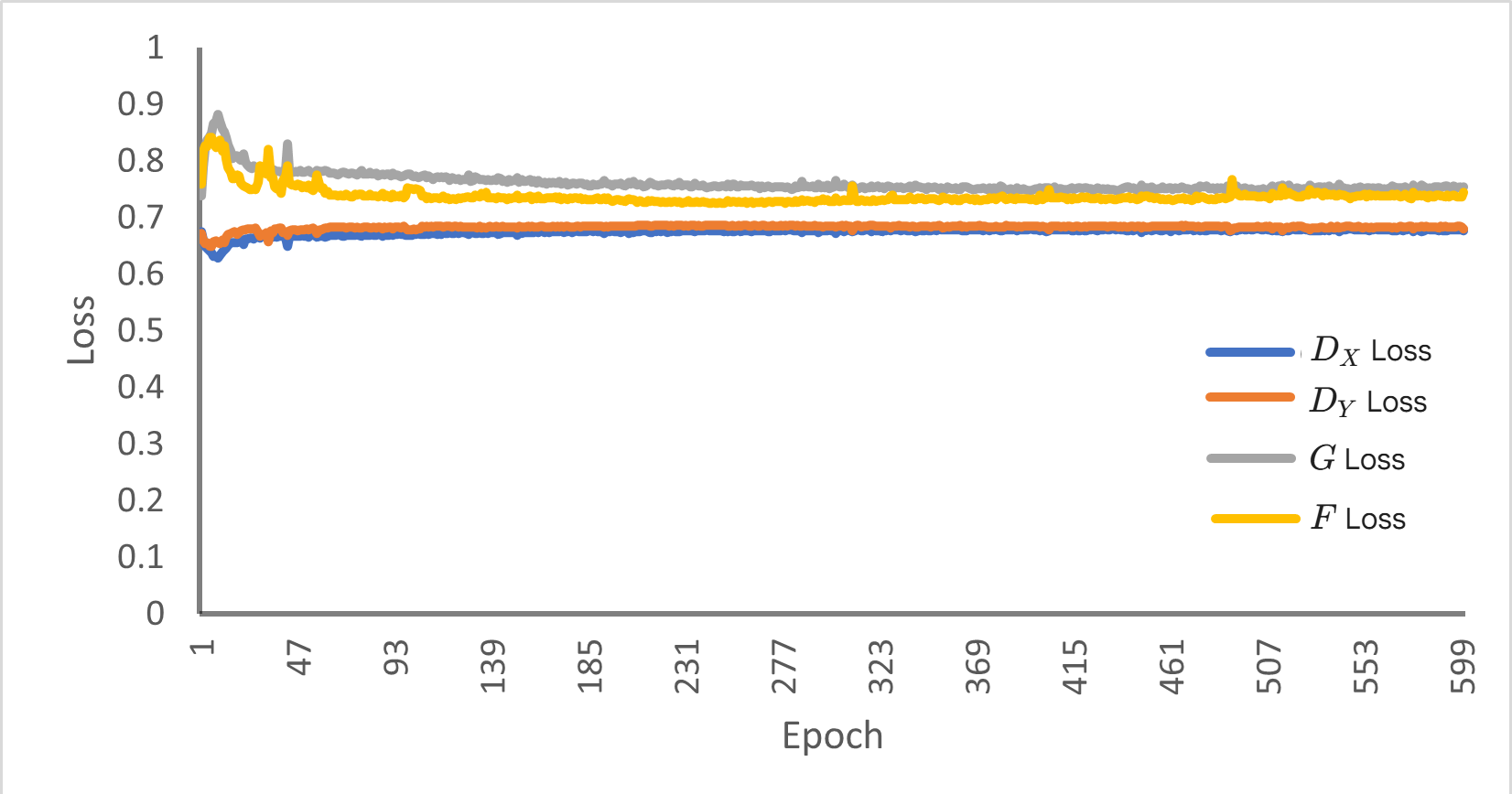}
\caption{ Convergence trajectories and training dynamics of the CycleGAN architecture. The plot illustrates the stable adversarial equilibrium achieved between the generators ($G, F$) and discriminators ($D_X, D_Y$). The convergence of discriminator losses toward the theoretical ideal ($\approx 0.69$) signifies a balanced competitive interplay, ensuring the synthesis of artifact-free dental morphology while preserving the structural integrity of the CBCT scans.}
\label{fig:trainingcurves}
\end{figure}

Figure \ref{fig:trainingcurves} delineates the convergence trajectories of the CycleGAN architecture, illustrating the iterative optimization of the generative and discriminative networks. The adversarial loss profiles demonstrate a reciprocal learning progression: the generators ($G$ and $F$) exhibit a steady refinement in synthesizing glare-reduced dental morphology, while the discriminators (${\mathbf{D}_X}$ and ${\mathbf{D}_Y}$) maintain a competitive threshold for authenticity. The absence of significant oscillations or divergence in these curves signifies a stable adversarial equilibrium, confirming that the model successfully minimized artifacts without sacrificing the underlying structural features of the CBCT scans. Further, the stability of the loss curves after the initial epochs indicates that the learning rate and hyperparameter configuration were optimal for the dental CBCT domain.

Beyond the individual network trajectories, the composite generator objective comprising a weighted summation of adversarial, cycle-consistency, and identity loss components was monitored to evaluate the global convergence of the architecture. This aggregate objective function exhibited an initial phase of rapid stochastic optimization, followed by an asymptotic stabilization. This trend signifies the successful attainment of a multi-objective equilibrium, wherein the model effectively decoupled metal-induced artifacts from the underlying dental morphology while ensuring the bidirectional consistency of the image-to-image translation.

\begin{figure}[H]
\centering
\includegraphics{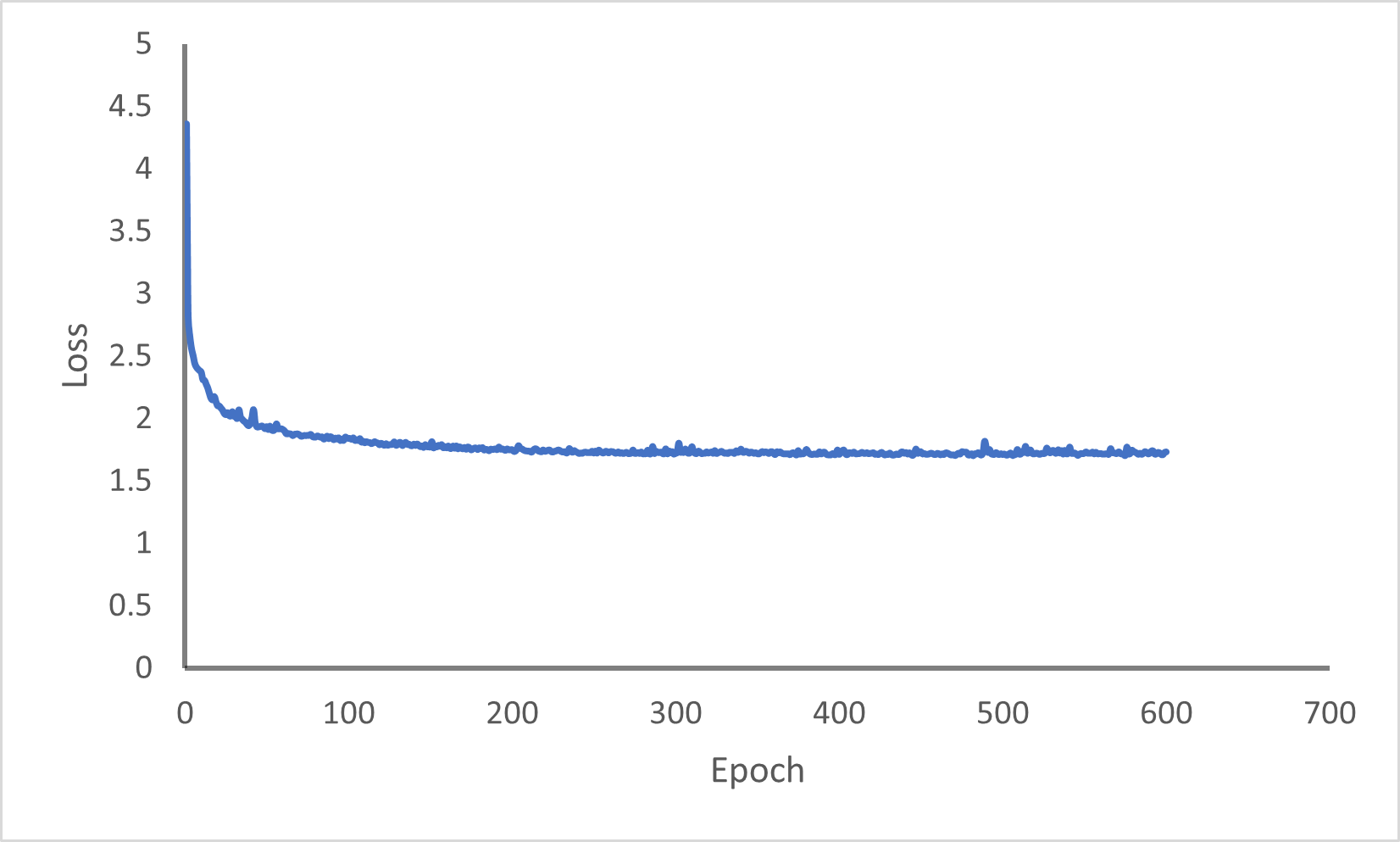}
\caption{Total generator objective trajectory. The curve shows the global optimization profile over 600 epochs. Stable asymptotic convergence confirms effective multi-objective loss minimization, successfully balancing adversarial realism with anatomical consistency.}
\label{fig:totgeneratorobjective}
\end{figure}

\begin{figure}[H]
\centering
\includegraphics{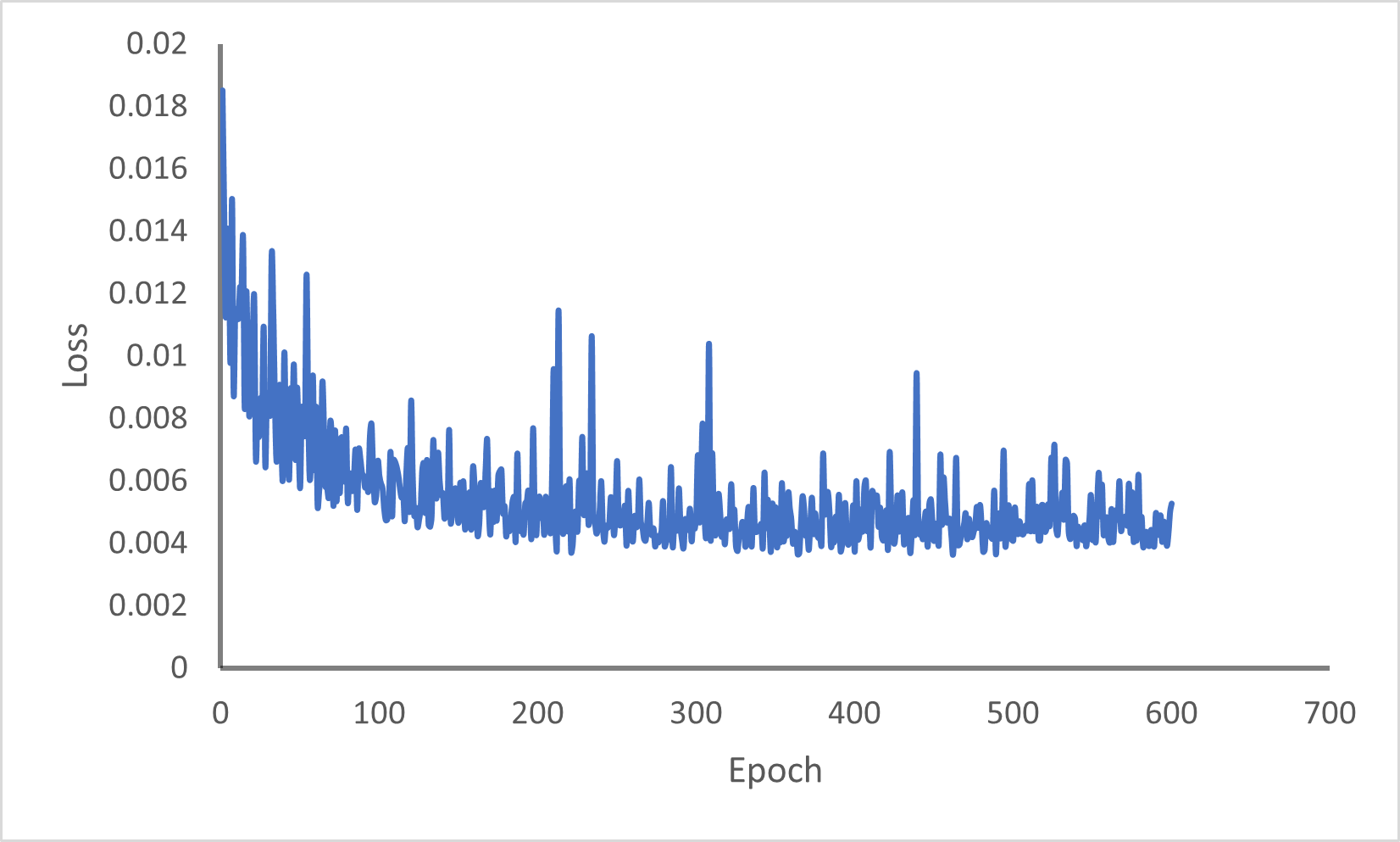}
\caption{Validation cycle-consistency loss. The steady decline and convergence of the cycle loss verify the structural integrity of the translation process, ensuring that dental morphology is preserved throughout the artifact reduction cycle.}
\label{fig:validate}
\end{figure}

To evaluate the model's generalization capacity and safeguard against overfitting, validation was conducted using the cycle-consistency loss (${\mathcal{L}_\mathrm{cyc}}$) on a held-out dataset. During this phase, unpaired volumes from both domains underwent reciprocal translation, and the $L_{{\mathrm{1}}}$ reconstruction error between the source images and their cyclic counterparts was quantified. The validation ${\mathcal{L}_\mathrm{cyc}}$ trajectory mirrored the training dynamics, exhibiting an asymptotic stabilization that confirms the model's ability to preserve structural integrity on previously unseen data. The convergence of both the composite generator objective (Figure \ref{fig:totgeneratorobjective}) and the validation cycle-consistency loss (Figure \ref{fig:validate}) served as the empirical basis for selecting the epoch 600 parameters, ensuring an optimal balance between artifact suppression and morphological constancy for final inference.

\subsection{Baseline Selection and Evaluation Framework}
To evaluate the performance of the proposed unsupervised CycleGAN framework, a multi-tiered benchmarking strategy was implemented, spanning classical deterministic algorithms and contemporary supervised deep learning architectures. The selection criteria aimed to quantify the efficacy of our unpaired approach against established clinical standards and paired deep-learning benchmarks.
\begin{itemize}
    \item \textbf{Classical Deterministic Baseline:} The Normalized Metal Artifact Reduction (NMAR) algorithm \cite{ref14} was selected as the representative clinical standard. NMAR remains a primary benchmark due to its stability in suppressing streak artifacts through sinogram interpolation, though it often lacks the structural preservation capabilities of generative models.
    \item \textbf{Supervised Deep Learning Benchmarks:} Two supervised variants—UNet2D1D and UNet2D2D \cite{ref34}—were integrated to serve as performance upper bounds. These architectures represent the modern standard for encoder-decoder based image restoration. By comparing our unsupervised framework against these supervised models, we evaluate the trade-offs between synthetic-pair training (which can introduce domain gaps) and our direct clinical-distribution learning strategy.
\end{itemize}

To ensure a fair and scientifically valid comparison, all evaluated methods were maintained on the following common grounds:

\begin{itemize}
    \item \textbf{Identical Testing Dataset:} All models were evaluated on the same held-out test set consisting of $n \approx 600$ axial slices. As detailed in Section 3.3, this partition was performed at the patient level to strictly prevent data leakage and ensure the assessment of model generalizability on unseen anatomy.
    \item \textbf{Standardized Preprocessing:} Every input image underwent identical Min-Max intensity normalization and was rescaled to a uniform resolution of $256 \times 256$ pixels to maintain architectural consistency across all benchmarks.
\end{itemize}

\subsubsection*{Evaluation Metrics and Rationale}

A hybrid evaluation framework was established to quantify restorative efficacy and anatomical integrity, specifically addressing the challenge of missing clinical ground truth.

\begin{enumerate}
    \item \textbf{Structural Fidelity (PSNR and SSIM):} While eak Signal-to-Noise Ratio (${\mathrm{PSNR}}$) and Structural Similarity Index Measure ${\mathrm{SSIM}}$ are traditionally full-reference metrics, they were strategically utilized here to quantify the \textbf{anatomical consistency} of the model's output relative to the original artifact-affected inputs. This approach ensures that the generative process preserves the patient's underlying dental morphology (e.g., pulp chambers and root structures) without introducing structural hallucinations or deformations. In this context, a higher ${\mathrm{SSIM}}$ indicates superior preservation of the anatomical ``identity'' during artifact suppression.
    
    \item \textbf{Perceptual Quality (BRISQUE):} To assess the naturalness of the reconstructed images without requiring a reference, the \textbf{Blind/Referenceless Image Spatial Quality Evaluator (BRISQUE)} was employed. This no-reference metric provides a standardized measure of perceptual quality, where lower scores indicate a reduction in artificial distortions and ``checkerboard'' artifacts.
    
    \item \textbf{Distributional Consistency (FID):} The \textbf{Fr\'{e}chet Inception Distance (FID)} was used to evaluate the similarity between the feature-space distributions of the generated results and a validated cohort of artifact-free clinical scans. A lower ${\mathrm{FID}}$ score validates the model's ability to shift the artifact-affected images into the domain of high-quality, ``clean'' dental CBCT imaging.
\end{enumerate}

\subsection{Ablation Study}

To verify the optimality of the proposed architecture and the selected loss weight combination (${\lambda_{\mathrm{cycle}}}=18$, ${\lambda_{\mathrm{id}}}=15$), a systematic ablation study was conducted. We evaluated the impact of key structural and functional components, specifically the identity loss (${\mathcal{L}_{\mathrm{id}}}$), the U-Net skip connections, and the sensitivity of the cycle-consistency weighting. The results are summarized in Table~\ref{table:ablation}.


\begin{table}[h!]
\centering
\caption{Results of the ablation study across different model configurations.}
\label{table:ablation}

\begin{tabular}{lcccc}
\hline
\textbf{Configuration} & \textbf{BRISQUE ($\downarrow$)} & \textbf{FID ($\downarrow$)} & \textbf{PSNR ($\uparrow$)$^{\ast}$} & \textbf{SSIM ($\uparrow$)$^{\ast}$} \\ \hline
\textbf{Full Model (Proposed)} & \textbf{10.82} & \textbf{157.04} & \textbf{25.96} & \textbf{0.9105} \\
\makecell[l]{Suboptimal Weights \\ ($\lambda_{\mathrm{cycle}}=10, \lambda_{\mathrm{id}}=5$)} & 12.95 & 168.43 & 23.18 & 0.8714 \\
Without Identity Loss (${\mathcal{L}_{\mathrm{id}}}$) & 13.67 & 174.21 & 22.45 & 0.8520 \\
Without U-Net Skip Connections & 14.32 & 181.57 & 21.05 & 0.8211 \\ \hline
\multicolumn{5}{l}{$^{\ast}$\small \textit{Calculated relative to the artifact-affected input to quantify anatomical structural fidelity.}}
\end{tabular}

\end{table}

As demonstrated in Table \ref{table:ablation}, the proposed full model configuration achieved the superior balance between artifact suppression and anatomical preservation. Notably, the removal of U-Net skip connections caused the most significant degradation in structural quality, with the ${\mathrm{SSIM}}$ dropping to $0.8211$ and ${\mathrm{PSNR}}$ to $21.05$ dB. This confirms that skip connections are essential for preserving high-frequency anatomical details during the translation process. Furthermore, the suboptimal weight configuration resulted in higher ${\mathrm{BRISQUE}}$ and ${\mathrm{FID}}$ scores, justifying the empirical selection of our final hyperparameter set and ensuring the model remains robust against generative hallucinations.

Overall, the results demonstrate that each component contributes significantly to the final performance. In particular, the combination of cycle-consistency and identity loss, together with the U-Net architecture, was found to be critical for achieving a balance between artifact removal and preservation of anatomical structures.

\subsection{Quantitative and Qualitative Evaluation of Reconstruction Accuracy}

As summarized in Table~\ref{table:comparison_results}, the proposed CycleGAN framework demonstrates superior performance across both perceptual and structural metrics compared to classical and supervised benchmarks. 

The classical NMAR algorithm\cite{ref14} yielded only a marginal improvement in distributional similarity, with the ${\mathrm{FID}}$ score decreasing slightly from $207.03$ to $202.68$. However, it resulted in a significant degradation of the perceptual quality, where the ${\mathrm{BRISQUE}}$ score increased to $25.43 \pm 6.25$. This divergence indicates that while NMAR reduces global high-intensity streaks through mathematical interpolation, it introduces severe unnatural blurring that disrupts local image statistics and diagnostic clarity.


Notably, the supervised deep learning baselines (UNet2D1D and UNet2D2D) \cite{ref34} failed to achieve clinically acceptable restorative results in this context, yielding high ${\mathrm{BRISQUE}}$  scores (exceeding 60) and ${\mathrm{FID}}$ values above 240. This performance degradation is attributed to the inherent domain gap between synthetic training pairs and real-world clinical distributions. Supervised architectures typically require voxel-perfect alignment between 'clean' and 'corrupted' images; in the absence of such ground truth, these models struggle to generalize to complex dental geometries, frequently introducing significant structural hallucinations and high-frequency noise that obscure diagnostic detail.

In contrast, the Proposed CycleGAN achieved a marked and consistent reduction in artifact-induced degradation. The model attained a ${\mathrm{BRISQUE}}$ score of $10.82 \pm 4.83$, representing a $34.6\%$ improvement over the artifact-affected baseline ($16.54 \pm 6.87$). Furthermore, the framework reached the lowest ${\mathrm{FID}}$ score of $157.04$, confirming its ability to successfully shift the corrupted image features into the domain of high-quality, artifact-free clinical scans.

Regarding structural preservation, the proposed model achieved an ${\mathrm{SSIM}}$ of $0.9105$ and a ${\mathrm{PSNR}}$ of $25.96$ dB. While NMAR yielded a higher ${\mathrm{PSNR}}$ ($29.96$ dB) due to its pixel-wise similarity to the blurred input, the CycleGAN's superior ${\mathrm{SSIM}}$ validates that the generative process preserves the structural identity and fine morphological details of dental tissues (e.g., pulp chambers and root canals) far more effectively than interpolation-based or supervised methods. Collectively, these findings validate that the unsupervised CycleGAN architecture facilitates a superior trade-off between aggressive artifact suppression and the high morphological preservation required for maxillofacial surgical planning.

\begin{table}[h!]
\centering
\caption{{Quantitative comparison of Metal Artifact Reduction (MAR) methods across different benchmarks.}}
\label{table:comparison_results}
\begin{tabular}{lcccc}
\hline
\textbf{Method} & \textbf{BRISQUE ($\downarrow$)} & \textbf{FID ($\downarrow$)} & \textbf{PSNR ($\uparrow$)$^{\ast}$} & \textbf{SSIM ($\uparrow$)$^{\ast}$} \\ \hline
Baseline (Artifact-Affected) & $16.54 \pm 6.87$ & 207.03 & - & - \\
NMAR (Meyer et al.) \cite{ref14} & $25.43 \pm 6.25$ & 202.68 & \textbf{29.96} & 0.8802 \\
UNet2D1D (Supervised)\cite{ref34} & $62.08 \pm 4.45$ & 263.28 & 16.64 & 0.5611 \\
UNet2D2D (Supervised)\cite{ref34} & $62.75 \pm 5.18$ & 248.77 & 16.32 & 0.5480 \\
\textbf{Proposed CycleGAN} & $\mathbf{10.82 \pm 4.83}$ & $\mathbf{157.04}$ & \textbf{25.96} & \textbf{0.9105} \\ \hline
\multicolumn{5}{l}{$^{\ast}$\small \textit{Calculated relative to the artifact-affected input to quantify anatomical structural fidelity.}}
\end{tabular}
\end{table}

The quantitative improvements are confirmed by the qualitative results. While the NMAR baseline reduces streak intensity, it often introduces blurring in the bone areas. In contrast, the proposed CycleGAN successfully removes metal streaks while sharpening the edges of dental structures. Notably, the preservation of fine details—specifically the periodontal ligament space and root canal morphology—demonstrates the model’s high accuracy. This shows that the model effectively distinguishes between metal noise and real anatomy, ensuring the images remain reliable for clinical use.

\begin{figure}[H]
\centering
\includegraphics[width=12cm]{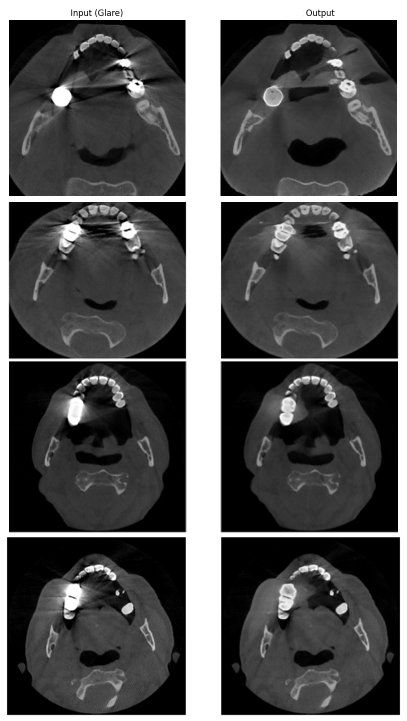}
\caption{Qualitative visual comparison between the artifact-affected input (left) and the proposed CycleGAN output (right).}
\label{fig:results_comparison}
\end{figure}

\subsection{ Computational Efficiency and Inference Time}

To evaluate the clinical feasibility of the proposed framework, we measured the inference time (runtime per image) for all models using an NVIDIA RTX 4050 GPU with a batch size of 1. As summarized in Table~\ref{table:runtime}, the proposed CycleGAN generator achieved the highest computational efficiency.

\begin{table}[h!]
\centering
\caption{Comparison of inference time per slice (Batch Size = 1) across different methods.}
\label{table:runtime}
\begin{tabular}{lc}
\hline
\textbf{Method} & \textbf{Inference Time (ms)} \\ \hline
UNet2D1D & $3.22 \pm 0.13$ \\
UNet2D2D & $4.17 \pm 0.12$ \\
\textbf{Proposed CycleGAN (Generator)} & $\mathbf{3.03 \pm 0.05}$ \\ \hline
\end{tabular}
\end{table}

The proposed model processed a single $256 \times 256$ slice in approximately 3.03 ms, which is significantly faster than the supervised UNet2D2D benchmark (4.17 ms). This efficiency stems from the optimized 2D convolutional architecture of the CycleGAN generator, making it highly suitable for real-time artifact reduction in clinical CBCT workflows.

\section{Discussion}

This study demonstrates the critical limitations of classical and supervised learning-based MAR techniques in dental CBCT and validates the superiority of an unsupervised domain-specific adversarial framework. 

While classical algorithms such as NMAR\cite{ref14} provide mathematically deterministic corrections, they are fundamentally limited by the scale of artifacts in dental imaging. Metallic implants create massive regions of photon starvation, resulting in extensive missing data. As evidenced by our quantitative results, NMAR's reliance on interpolation leads to severe over-smoothing of high-frequency details, reflected in its worsened ${\mathrm{BRISQUE}}$ score of $25.43 \pm 6.25$.  The elevated FID scores for supervised benchmarks  (UNet variants) reflect a lack of distributional alignment with real-world dental anatomy when trained on simulated metal artifacts. Furthermore, the failure of supervised benchmarks highlights the critical ``domain gap'' in clinical settings. Without voxel-aligned paired data, supervised models introduce significant structural hallucinations, as indicated by their low ${\mathrm{SSIM}}$ scores ($\approx 0.55$), making them unreliable for diagnostic tasks.

The proposed unsupervised CycleGAN framework transcends these limits. By utilizing a U-Net generator with skip connections\cite{ref10}, the model does not merely ``bridge gaps''; rather, it learns the underlying morphological distribution of healthy dental anatomy. The preservation of anatomical identity is evidenced by the framework's superior ${\mathrm{SSIM}}$ of $0.9105$ and an ${\mathrm{FID}}$ score of $157.04$, confirming that the synthesized bone and dental textures are clinically realistic and statistically aligned with artifact-free clinical scans.
Furthermore, qualitative assessments by domain experts confirm that the framework predominantly preserves anatomical structural integrity; while minor generative hallucinations were observed in a small subset of samples with extreme data loss, the vast majority of outputs maintained high-fidelity dental morphology suitable for clinical observation."

\subsection{Qualitative Analysis of Failure Cases and Hallucinations}

\begin{figure}[H]
\centering
\includegraphics[width=12cm]{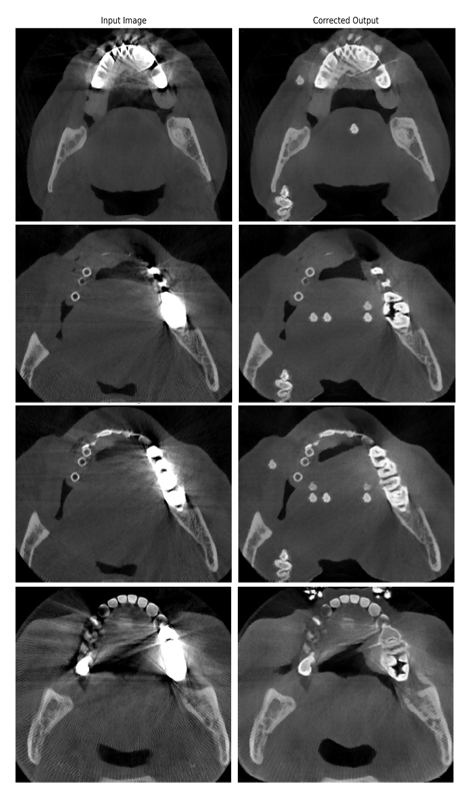}
\caption{{Qualitative analysis of representative failure cases and generative hallucinations.}}
\label{results fail}
\end{figure}

A rigorous evaluation identified specific instances where the model's performance reached its information-theoretic limits. Due to the generative nature of the CycleGAN architecture, the framework remains susceptible to 'hallucinations'—the stochastic synthesis of anatomically unverified features in regions of extreme data loss. As illustrated in Figure \ref{results fail}, scenarios involving multiple, closely-spaced high-density implants can result in 'photon starvation' so severe that original structural data is irrecoverable. In these isolated cases, the model may generate localized, repetitive textures that lack the intricate stochasticity of true trabecular bone patterns.

Due to these generative characteristics, we emphasize that the proposed framework is intended to serve as a supportive diagnostic aid rather than a standalone clinical tool. The final interpretation of the reconstructed CBCT volumes must always be validated by a clinical expert (e.g., a radiologist or maxillofacial surgeon) to ensure that the restored details are grounded in anatomical reality and free from generative artifacts. This human-in-the-loop approach is critical to prevent diagnostic errors arising from AI-induced hallucinations.

\subsection{Limitations of the Study and Future Work}

While the proposed unsupervised framework demonstrates significant restorative capabilities, we acknowledge several limitations that provide a roadmap for future research:

\begin{enumerate}
    \item \textbf{Lack of Clinical Ground Truth:} The most significant challenge in clinical MAR research is the absence of voxel-aligned, artifact-free references for real patients. Although we addressed this by using no-reference metrics (${\mathrm{BRISQUE}}$ and ${\mathrm{FID}}$) and structural fidelity indices (${\mathrm{SSIM}}$ relative to the input), these are proxy measures. Future studies should incorporate physical dental phantoms with known geometries to establish absolute error margins.
    
    \item \textbf{Generative Hallucinations and Clinical Oversight:} As discussed in Section 5.1, the generative nature of CycleGAN can occasionally introduce anatomical hallucinations in regions of extreme photon starvation. Therefore, the current model must be viewed as a supportive diagnostic tool. Clinical expert validation remains mandatory to distinguish between authentic anatomical restoration and potential AI-induced artifacts.
    
    \item \textbf{2D Spatial Context vs. 3D Volumetric Continuity:} Our current approach operates on 2D axial slices. While computationally efficient, this ignores the inter-slice dependency along the vertical ($z$) axis. Future work will focus on extending the unsupervised adversarial learning to 3D volumetric data or 2.5D architectures to enhance vertical structural consistency.
    
    \item \textbf{Model Generalizability Across CBCT Vendors:} The model was trained and validated using a specific dataset (ToothFairy) which reflects the reconstruction kernels and noise profiles of a particular subset of CBCT scanners. The robustness of the framework across different hardware vendors and scanning protocols (e.g., varying peak voltages or filtration) has not been fully evaluated. Future work will investigate domain adaptation techniques to ensure the model remains scanner-agnostic.

  \item \textbf{Diverse Metallic Objects and Prosthetics:} While the framework effectively reduces artifacts from common dental restorations, its performance on specialized high-density materials—such as titanium implants or orthodontic appliances—has not been fully evaluated. These objects create distinct "photon starvation" patterns that differ from standard fillings. To address this, future work will involve collecting and labeling local clinical data to expand the training distribution, ensuring the model generalizes across a broader range of metallic prosthetics and diverse patient populations. 
  
\end{enumerate}

\section{Conclusion}

This study presents an unsupervised framework based on CycleGAN for the mitigation of metal-induced artifacts in dental CBCT imaging. By leveraging an unpaired image-to-image translation approach, the proposed method bypasses the logistical and ethical constraints associated with acquiring large-scale, voxel-perfect labeled clinical datasets. 

Our findings demonstrate that the architecture effectively suppresses high-intensity artifacts while maintaining high anatomical fidelity --- a claim substantiated by a 34.6\% improvement in ${\mathrm{BRISQUE}}$ scores, a substantial reduction in \textbf{${\mathrm{FID}}$ ($207.03 \rightarrow 157.04$)}, and a superior ${\mathrm{SSIM}}$ of $0.9105$. Crucially, the proposed model significantly outperformed both the classical NMAR algorithm and supervised deep learning benchmarks, proving that unsupervised adversarial learning can successfully reconstruct viable bone morphology where traditional interpolation and supervised models fail due to domain gaps.

Beyond restorative accuracy, the framework demonstrates high clinical feasibility with a rapid average inference time of $3.03 \pm 0.05$ ms. This high-speed processing capability satisfies the requirements for real-time integration into digital dentistry workflows, allowing for near-instantaneous diagnostic feedback. While the generative nature of the model necessitates clinical expert oversight to mitigate potential hallucinations in extreme photon-starvation zones, the framework serves as a robust, supportive diagnostic aid. 

Future work will focus on extending this unsupervised approach to raw 3D volumetric projection data and implementing domain adaptation techniques to ensure the model remains scanner-agnostic across various hardware vendors. These advancements will further enhance the precision and accessibility of metal artifact reduction in maxillofacial radiology.

\section*{Author Contributions}
Conceptualization, M.D. and R.D.J.; methodology, M.D., P.H.S.V.N., S.N.A.D. and G.L.T.C.; software, P.H.S.V.N., S.N.A.D. and G.L.T.C.; validation, M.D. and R.D.J.; formal analysis, M.D.; investigation, P.H.S.V.N., S.N.A.D., G.L.T.C., M.D. and R.D.J.; resources, M.D. and R.D.J.; data curation, R.D.J., P.H.S.V.N., S.N.A.D. and G.L.T.C.; writing---original draft preparation, M.D., P.H.S.V.N., S.N.A.D. and G.L.T.C.; writing---review and editing, M.D.; visualization, P.H.S.V.N., S.N.A.D. and G.L.T.C.; supervision, M.D. and R.D.J.; project administration, M.D. All authors have read and agreed to the published version of the manuscript.

\section*{Funding}
This research received no external funding.

\section*{Institutional Review Statement}
Not applicable for this study (no human or animal subjects involved).

\section*{Data Availability Statement}
The processed data subset utilized in this study is a derivative of the publicly available ToothFairy dataset \cite{ref13} (\url{https://ditto.ing.unimore.it/toothfairy/}). The specific curated 2D slices and partitions generated for this research are available for reproducibility purposes at \url{https://github.com/maheshi81/CBCT_MAR}. Original raw volumetric scans remain the property of the ToothFairy challenge organizers.

\section*{Acknowledgments}
We acknowledge the providers of the ToothFairy dataset, whose open-access repository was instrumental in the training and validation of the proposed generative models.
We also acknowledge that AI tools were used during drafting to improve sentence structure, readability, and language. The content was reviewed and edited as needed, and the author is fully responsible for the publication's content.

\section*{Conflicts of Interest}
The authors declare no conflicts of interest.


\end{document}